\pdfoutput=1

\documentclass[11pt]{article}

\usepackage[]{acl}

\usepackage{times}
\usepackage{latexsym}

\usepackage[T1]{fontenc}

\usepackage[utf8]{inputenc}

\usepackage{microtype}

\usepackage{inconsolata}

\usepackage{hyperref}
\usepackage{url}

\usepackage{comment}
\usepackage{todonotes}
\usepackage{tabularx}
\usepackage{multirow}
\usepackage{graphicx}
\usepackage{booktabs}
\usepackage{arydshln}
\usepackage{pifont}
\usepackage{xcolor}
\usepackage{tcolorbox}
\usepackage{wrapfig}

\usepackage{xspace}

\newcommand{\ourmodel}{\texttt{ReEval}\xspace}

\input{Definitions}

\title{

\ourmodel: Automatic Hallucination Evaluation for Retrieval-Augmented Large Language Models via Transferable Adversarial Attacks}

\author{
Xiaodong Yu$^{\spadesuit}$\thanks{Work done during an internship at Microsoft Research.} \enspace\enspace\enspace Hao Cheng$^\clubsuit$ \enspace\enspace\enspace Xiaodong Liu$^\clubsuit$ \enspace\enspace\enspace Dan Roth$^\spadesuit$ \enspace\enspace\enspace Jianfeng Gao$^\clubsuit$ \\
$^\spadesuit$University of Pennsylvania \enspace\enspace\enspace $^\clubsuit$Microsoft Research\\
\url{https://autodebug-llm.github.io/}
}

\begin{document}

\maketitle

\begin{abstract}

Despite remarkable advancements in mitigating hallucinations in large language models (LLMs) by retrieval augmentation, it remains challenging 
to measure the reliability of LLMs using \textit{static} question-answering (QA) data.
Specifically, given the potential of data contamination (\eg leading to memorization), good static benchmark performance does not ensure that model can reliably use the provided evidence for responding, which is essential to avoid hallucination when the required knowledge is new or private. 
Inspired by adversarial machine learning, we investigate the feasibility of automatically perturbing existing static one for \textit{dynamic} evaluation.
Specifically, this paper presents \ourmodel, an LLM-based framework using prompt chaining to perturb the original evidence for generating new test cases for evaluating the LLMs' reliability in using new evidence for answering. 

We implement \ourmodel using ChatGPT and evaluate the resulting variants of two popular open-domain QA datasets on a collection of LLMs under various prompting settings.
Our generated data is human-readable and useful to trigger hallucination in LLM.
Accurate models on static data are observed to produce unsupported answers from the perturbed evidence, with pronounced accuracy drops across LLMs including GPT-4.
We find that our adversarial examples are transferable across all considered LLMs. The examples generated by a small model can be used to evaluate a much larger model, making our approach cost-effective.

\end{abstract}
\section{Introduction}
\label{sec:intro}
Due to their superior capability in generating coherent and convincing outputs, large language models (LLMs), such as ChatGPT \citep{openai2022chatgpt}, GPT4 \citep{openai2023gpt4}, Claude \citep{claude2} and Palm \citep{anil2023palm}, have been extensively used as foundations for language technologies.


Though LLMs excel in memorizing knowledge and understanding natural language, merely depending on parametric knowledge for inquires (closed-book) has inherent limitations. 
Specifically, these models are unaware of knowledge update and uninformed about new or private information they have not previously encountered. 
One popular way to mitigate this is to augment LLMs with external relevant evidence (open-book), \eg retrieval-augmented LLMs \citep{shi2023replug,peng2023check}, outperforming their closed-book counterparts.
However, this improvement does not necessarily imply that the model with retrieval augmentation \textit{truly integrates the given evidence for deriving the response}.
As most popular datasets used for evaluation are curated using public corpora (\eg Wikipedia), which are already included in the LLM pretraining, they risk becoming not challenging enough, and models may achieve higher accuracy by mere memorization or by exploiting their familiarity with topics or domains found in static evaluation datasets.
Thus, it raises concerns as to whether retrieval-augmented LLMs might resort to fabricating answers that are inconsistent with the presented evidence, resulting in hallucination. 
Given the wide applications of retrieval-augmented LLMs, it is critical to reliably assess their faithfulness to the context for trustworthy and safe AI, particularly when handling sensitive or recently updated information.


\begin{figure*}[t!]
    \centering
    \includegraphics[width=0.8\textwidth]{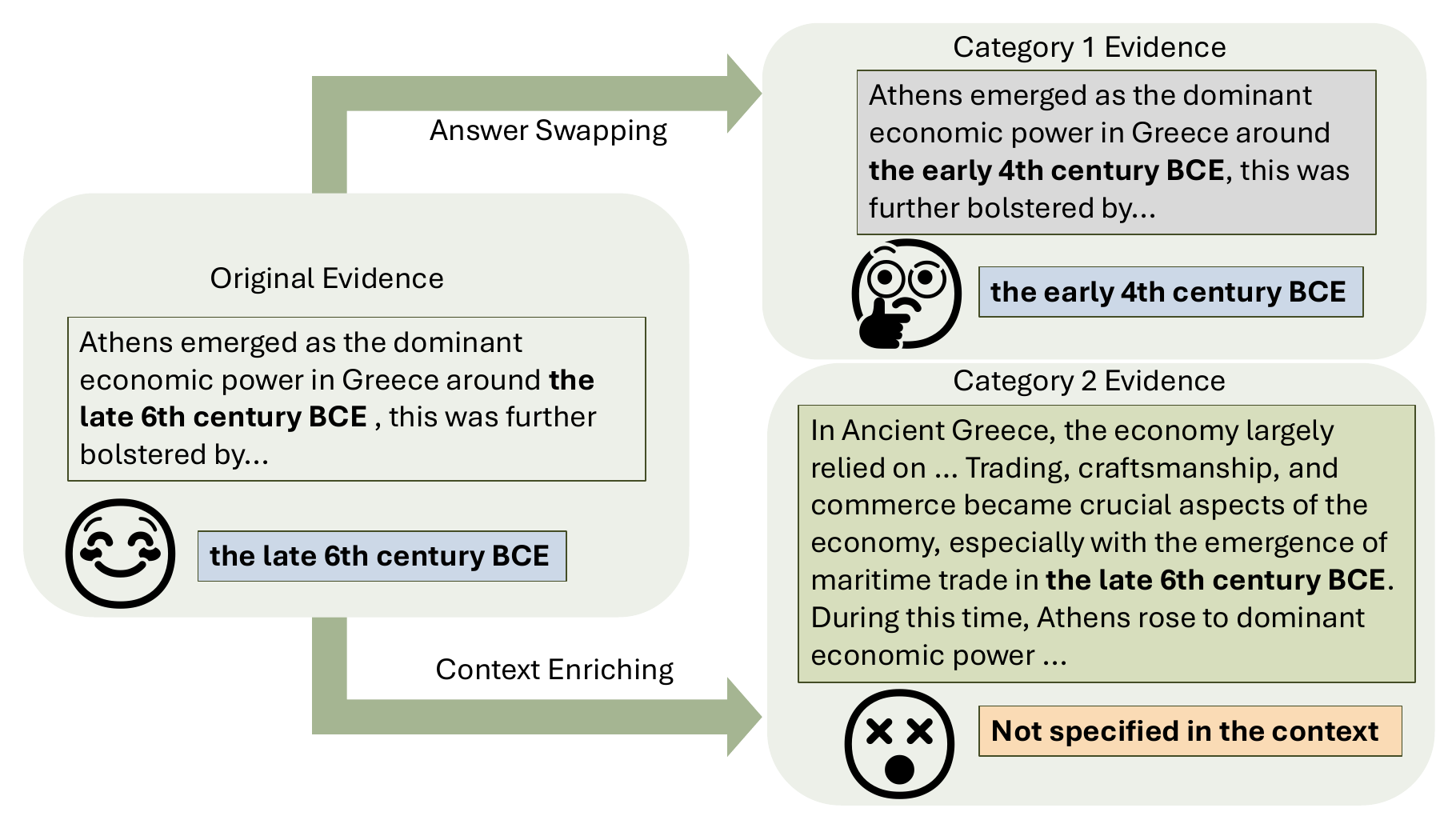}
    \caption{An example of how the original evidence is edited (answer swapping and context enriching) by \ourmodel. The question is ``when did athens emerges as wealthiest greek city state?". ``the early 4th century BCE'' and ``the late 6th century BCE''is the desirable answers for \textit{answer swapping} (Category 1) and \textit{context enriching} 
 (Category 2), respectively. ChatGPT answers are next to the emoji.
    }
    \label{fig:intro_example}
\end{figure*}

In this work, we propose a new evaluation framework \ourmodel, which dynamically generate new data to evaluate LLMs.
Motivated by using adversarial attacks to trigger undesirable behaviors in machine learning models \citep{madry2017pgd,goodfellow2014explaining}, we focus on perturbing evidence in the prompts to measure the reliability of LLMs' capability of deriving proper responses based on the provided context.
Through the perturbation of either the answer span or the rest context in the given evidence, \ourmodel accordingly provides two ways of synthesizing evaluation datasets (see examples in \autoref{fig:intro_example}): 
1) \textit{answer swapping} (Category 1), where the original answer is replaced with another valid answer while the remaining context is intact;
2) \textit{context enriching} (Category 2), where more relevant information is added to the provided document while the original supportive information is kept.
The former simulates the scenario where only the answer-relevant part of the document is updated while the latter represents the evolving document where more related information is added leading to more complex documentation of specific topics. We then implement \ourmodel by \textit{prompt chaining} with LLMs, \ie using LLMs to generate new test cases that are more likely to trigger hallucinations in LLMs.


To verify the effectiveness of the proposed framework, we apply it to two popular open-domain QA dataset, Natural Questions (NQ) \citep{kwiatkowski-etal-2019-natural} and RealtimeQA \citep{kasai2022realtime}.
Human studies are conducted to verify the naturalness of the generated adversarial attacks, \ie the updated document is human-readable, supporting the \textit{desirable} answer for the corresponding question. We then evaluate our generated datasets on both open-source (Alpaca \citep{alpaca}) and propriety (ChatGPT, Claude, Palm and GPT-4) LLMs under various prompting settings, \eg zero-shot, few-shot, and more enhanced prompting techniques designed to improve the reliability of prompting with LLMs.
Although natural and supportive in the eyes of humans, both probing datasets trigger LLMs to produce inconsistent answers based on the perturbed evidence, regardless of their model sizes and training techniques.
We find that the self-attacks are more effective but attacking test examples generated by our method is transferable across all considered LLMs.
This enables the possibility of evaluating LLMs using test cases generated by more cost-effective LLMs.
\begin{figure*}[t!]
    \centering
    \includegraphics[width=\textwidth]{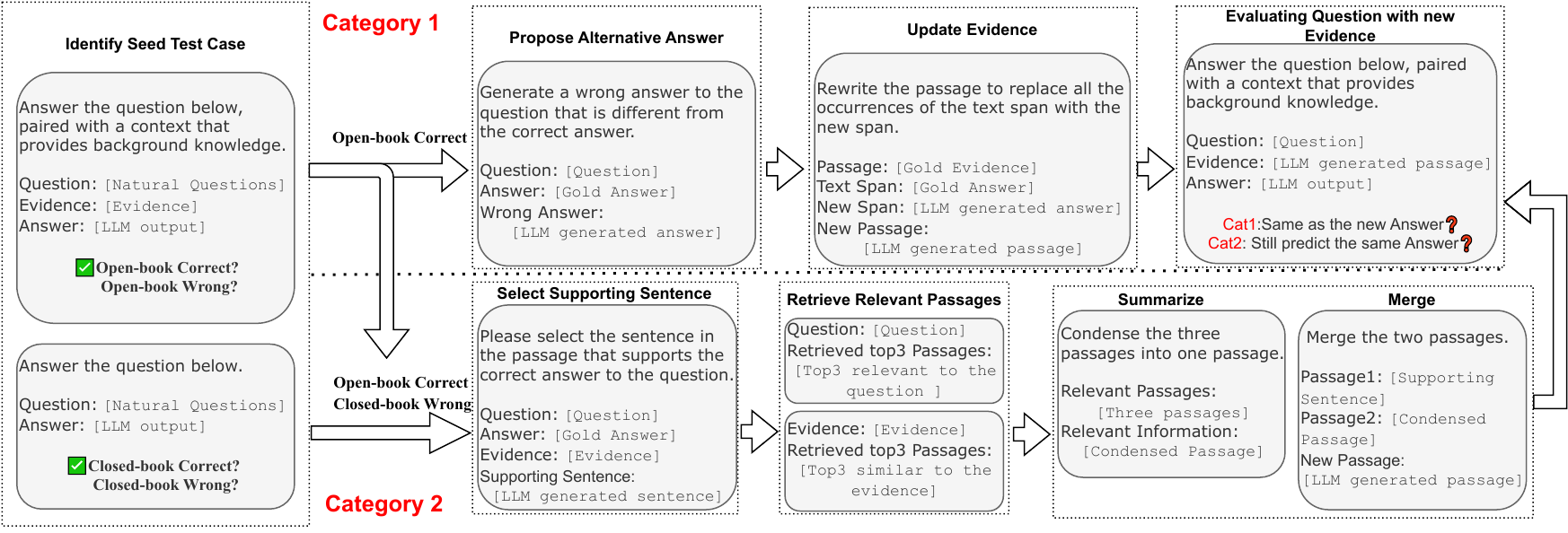}
    \caption{The pipeline of \ourmodel, including identifying seed cases, generating new tests, and hallucination evaluation.}
    \label{fig:pipeline}
\end{figure*}

\section{Related Work}
\label{sec:related_work}

\paragraph{Faithfulness of Augmented LLM.} Recent work shows that, given the correct passages, LLMs could be highly receptive to the provided passage even if the passage is inconsistent with the model memory.
For example, \cite{xie2023adaptive} focus on machine-generated questions from a subject-object-relation triple with machine-generated evidence, and
\cite{zhou2023context} design prompt templates that could force the model to follow the provided context and thus improve the faithfulness of the model.
Instead, we use diverse and real-world questions from NQ and focus on editing the passage without compromising the naturalness of the original passages.
In addition to including the advanced prompting from \cite{zhou2023context} in our study, we focus on a more diverse and challenging set of questions rather than a smaller and simpler one with questions that could be answered correctly under the zero-shot closed-book setting.
We argue that the difficulty and diversity of the questions as well as the naturalness of evidence passages are crucial for understanding the hallucination of SOTA LLMs for real-world applications.
In our framework, we keep the questions natural, and the evidence is from Wikipedia with abundant information.
For Category 1 data generation, previous work introduces ideas on altering the entities in the passage \citep{yan2021robustness, longpre2021entity, zhou2023context}, while we consider all types of answers (including entities as subcases), and use LLM to automatically substitute the answer properly (making it fit the context).
For Category 2 data generation, \cite{choi2021decontextualization} propose to decontextualize the supporting sentence from the passage, and \cite{jia-liang-2017-adversarial} add distractors to the original passage. In contrast, we want to enrich the original passage by first extracting the supporting sentence with proper decontextualization and then enriching it with other relevant information based on prompting with LLMs.

\paragraph{Adversarial Attacks \& Transferability.}
There is a long line of research in generating adversarial examples to trigger errors or undesirable behaviors from machine learning models \citep{szegedy2014intriguing,goodfellow2014explaining}.
To improve the robustness of machine learning models, there are also a number of methods proposed to defend against such attacks \citep{madry2017pgd,zhu2019freelb,li2020textat,cheng-etal-2021-posterior}.
However, models trained with adversarial learning are found to have at-odds generalization \cite{tsipras2018robustness,zhang2019theoretically}, \eg improving the accuracy on adversarial attacks can compromise the model performance on clean examples. 
Despite being more challenging due to its discrete nature, different text adversarial attacks with perturbed inputs imperceptible to humans have been proposed for question answering \citep{jia-liang-2017-adversarial}, natural language inference \citep{nie-etal-2020-adversarial}, and sentiment classification \citep{iyyer-etal-2018-adversarial}.
One surprising phenomenon is that many adversarial examples are \textit{transferable} \citep{papernot2016transferability,wallace2021universal}. For example, \citet{wallace2021universal} show that adversarial prefix optimized for one particular model can also transfer to models of different architectures and sizes.
In addition to relying on white-box access to generate effective adversarial examples, recent work even reports that it is difficult to generate reliable examples via automatic search \citep{carlini2023aligned}.
Our work is highly motivated by this long line of work, \ie making evidence edits while keeping the input legitimate for the targeted task so that the LLMs cannot reliably answer the question.
Here, we do not assume any model access except its text outputs, \ie black-box.
We show that our proposed approach of generating adversarial test cases from a pivot LLM can trigger hallucination behaviors across a set of open-source and proprietary LLMs.

\section{\ourmodel Framework}
\label{sec:method}

Assessing the hallucination of LLMs is challenging as we often do not know what changes in the prompt would trigger LLMs to hallucinate.
In this paper, we present our approach \ourmodel\ for automatically constructing a large number of test cases that can surface hallucination issues.
Given a pivot LLM, we first prompt it to identify \textit{seed test cases} from a pool of existing data.
Then we prompt the pivot LLM again to generate \textit{attacking test cases} based on individual seed test cases.
These attacking test cases are used to evaluate the performance of the pivot LLM (\textbf{self-attack}) as well as other LLMs (\textbf{cross-attack}).
While \ourmodel\ is a general framework, we focus on the QA scenario where the LLMs to be evaluated need to answer open-domain questions based on their supporting evidence.
The pipeline is illustrated in \autoref{fig:pipeline}.

\paragraph{Seed Case Selection.} To identify seed test cases, we categorize QA examples 
based on whether the pivot LLM can answer the question correctly under
the open-book and closed-book settings in a zero-shot fashion, similar to typical static evaluation.
In the closed-book setting, only the question itself is given and the pivot LLM can only rely on memorization,
whereas in the open-book setting, the associated supporting evidence is provided.
As we are interested in assessing \textbf{whether the LLM can truely comprehend the provided evidence and reliably use that for answering}, only cases that can be answered correctly using open-book prompt are kept as seed.
For those cases, \ourmodel generates attacking test cases by perturbing the evidence, potentially updating the answers (\eg answer swapping).
Below is the zero-shot open-book prompt for seed test case selection,
and the closed-book version simply drops the evidence part (see more examples in Appendix).

\begin{tcolorbox}[title=\footnotesize Zero-shot Open-book Prompt,top=1mm,bottom=1mm]
\scriptsize
Answer the question below, paired with a context that provides background knowledge. Only output the answer without other context words.

Context: \{Evidence\}

Question: \{Question\}

Answer: 
\end{tcolorbox}

\paragraph{Evidence Perturbation.}
To generate viable attacking test cases, we consider the following two perturbation approaches.
\begin{enumerate}
    \item \textbf{Answer Swapping} (top flow in \autoref{fig:pipeline}): Update the evidence using a new answer that may lead to a knowledge conflict (\S\ref{ssec:cat1}). 
    In the top-right example of \autoref{fig:intro_example}, we replace 
    {\it ``the late 6th century BCE''} with {\it ``the early 4th century BCE''} in the evidence
    and test whether the LLM can update its answer accordingly.
    
    \item \textbf{Context Enriching} (bottom in \autoref{fig:pipeline}): Enrich the evidence using extra relevant facts that may dilute the information (\S\ref{ssec:cat2}).
    In the bottom-right example of \autoref{fig:intro_example}, the evidence becomes much more dense
    though the answer is unchanged, and we test whether the LLM can still produce the original answer.
\end{enumerate}

For the second approach, we exclude cases where the pivot LLM can answer correctly under the closed-book setting
since perturbing the evidence for such cases may not surface the hallucination issue, 
\ie the LLM may simply use its internal memory to answer the question correctly and completely ignore the evidence. 

\paragraph{Re-evaluation.} To assess the hallucination of LLMs, we can simply measure the accuracy of
the predicted answers for the attacking test cases. 
If the LLM faithfully follows the provided context, it should be immune to these perturbations and maintain a high accuracy score.
The evaluation considers both zero-shot and few-shot prompting.
The zero-shot prompt for evaluation is identical to the one used for seed test selection above.
The few-shot version inserts the demonstrations of evidence-question-answer triplets right before the ``Context: \{Evidence\}'' line.

\begin{tcolorbox}[title=\footnotesize Few-shot Open-book Prompt,top=1mm,bottom=1mm]
\scriptsize
Answer the question below, paired with a context that provides background knowledge. Only output the answer without other context words.

{\{Demonstrations of Evidence-Question-Answer tuples\}}

Context: \{Evidence\}

Question: \{Question\}

Answer: 
\end{tcolorbox}

\subsection{Category 1: Answer Swapping}
\label{ssec:cat1}
Here, we present the first approach to generate test cases by updating the original evidence with alternative answers. Specifically, those alternative answers are proposed by the pivot LLM via prompting. Note that the considered seed test cases are open-book correct with the pivot LLM.

For each question, given the original answer and supportive evidence, we first ask the model to generate an alternative answer that is factually wrong using the following prompt.
\begin{tcolorbox}[title=\footnotesize Prompt for Generating An Alternative Answer,top=1mm,bottom=1mm]
\scriptsize
Generate a wrong answer to the question that is different from the correct answer.

Question: \{Question\}

Answer: \{Gold Answer\}

Wrong Answer:
\end{tcolorbox}
We then instruct the LLM to replace all the occurrences of the original answer with the alternative one.\footnote{Although a simple string match can also do the job, it can make the answer occurring sentences inconsistent with the neighboring context, \eg mismatched pronouns and aliases.}
\begin{tcolorbox}[title=\footnotesize Prompt for Updating Evidence,top=1mm,bottom=1mm]
\scriptsize
Rewrite the passage to replace all the occurrences of the text span with the new span.

Passage: \{Original Evidence\}

Text Span: \{Original Answer\}

New Span: \{LLM generated answer\}

New Passage:
\end{tcolorbox}
Since most context is kept, the newly generated evidence is likely to support the alternative answer for most questions (as verified in \S\ref{sec:mturk}).

\subsection{Category 2: Context Enriching}
\label{ssec:cat2}
Our second strategy aims to enrich the original evidence with more relevant context, leading to a more complex context for answer reasoning.
Unlike Category 1 discussed above, we only keep seed cases that are open-book correct but closed-book wrong to ensure that certain comprehension of the evidence is required to answer the question correctly.

To ensure that the newly generated evidence still provides support for the question, we first extract the supporting sentence from the original evidence.
\begin{tcolorbox}[title=\footnotesize Prompt for Selecting the Supporting Sentence,top=1mm,bottom=1mm]
\scriptsize
Please select the sentence in the passage that supports the correct answer to the question.

Question: \{Question\}

Answer: \{Answer\}

Evidence: \{Evidence\}

Supporting Sentence:
\end{tcolorbox}
We then gather relevant information from an external database to be used for composing the new evidence.
Here, we consider two ways of retrieving passages from Wikipedia for fusion with the supporting sentence above,
\ie evidence-focused expansion and question-focused expansion, where the former uses the original evidence as the query and the question is used for the latter case.
As these two expansions bring in different types of relevant information, we create two corresponding copies of new evidence.
To make the information more diverse, we select the top-$k$ passages from different Wikipedia pages.
To merge these passages into a single passage, we first ask the LLM to summarize the information of the retrieved set,
and then merge the supporting sentence into the summary.
Here, the pivot LLM needs to extract and summarize key information 
so that the new evidence is human-readable and still supports the original answer.

\tcbset{width=(0.93\linewidth)/2,equal height group=AT,before=,after=\hfill,top=1mm,bottom=1mm}
\begin{tcolorbox}[title=\footnotesize Summarize Prompt]
\scriptsize
Condense the three passages into one passage.

Relevant Passages: \{List of Passages\}

Relevant Information:
\end{tcolorbox}
\begin{tcolorbox}[title=\footnotesize Merge Prompt]
\scriptsize
Merge the two passages

Passage1: \{Supporting Sentence\}

Passage2: \{Condensed Passage\}

New Passage:
\end{tcolorbox}

\begin{table*}[t!]
        \small
        \setlength{\tabcolsep}{5pt}
        \centering
        \begin{tabular}{llccccccccc}
            \toprule
            \multirow{2}{*}{Models} & \multirow{2}{*}{Method} & \multicolumn{4}{c}{Zero-shot} & & \multicolumn{4}{c}{Few-shot}\\ 
            \cline{3-6}
            \cline{8-11}
             &  & EM & F1 & Entail. & Norm Entail. & & EM & F1 & Entail. & Norm Entail.\\
            \midrule
            \multirow{2}{*}{Alpaca-7B} & Open-book & 18.71 & 36.04 & 56.65 & 71.68 && 21.50 & 38.46 & 57.30 & 67.45\\
            & Faithful Prompt & 27.80 & 43.64 & 58.75 & 68.86 && 33.74 & 51.10 & 65.41 & 74.33\\
            \midrule
            \multirow{2}{*}{ChatGPT} & Open-Book & 43.71 & 59.99 & 77.31 & 77.31 && 40.44 & 54.58 & 65.33 & 65.33\\
            & Faithful Prompt & 44.73 & 40.04 & 42.98 & 42.98 && 40.04 & 52.75 & 62.11 & 62.11\\
            \midrule
            \multirow{2}{*}{Claude 2} & Open-Book & 44.62 & 56.37 & 59.08 & - && 20.32 & 34.09 & 69.77 & -\\
            & Faithful Prompt & 52.95 & 65.05 & 71.80 & - && 39.28 & 50.97 & 71.83 & -\\
            \midrule
            \multirow{2}{*}{Palm} & Open-Book & 57.50 & 65.75 & 74.71 & 80.13 && 65.75 & 75.74 & 78.41 & 83.38\\
            & Faithful Prompt & 64.17 & 68.41 & 79.20 & 84.19 && 68.41 & 78.61 & 81.46 & 86.15\\
            \midrule
            \multirow{2}{*}{GPT-4} & Open-Book & 54.11 & 68.50 & 81.29 & 84.73 && 58.94 & 72.58 & 81.01 & 83.79\\
            & Faithful Prompt & 58.49 & 71.70 & 82.51 & 85.52 && 63.49 & 75.72 & 82.25 & 85.19\\
           
            \bottomrule
        \end{tabular}
        \caption{Zero-shot and few-shot performance of LLMs on Category 1 data of NQ. ``Entail.'' refers to the entailment accuracy. ``Norm Entail.'' refers to the entailment accuracy of the normalized test set that only includes the accurate cases before perturbation.}
        \label{tab:3.3_cat1_result}
    \end{table*}

\begin{table*}[t!]
        \small
        \setlength{\tabcolsep}{5pt}
        \centering
        \begin{tabular}{llccccccccc}
            \toprule
            \multirow{2}{*}{Models} & \multirow{2}{*}{Method} & \multicolumn{4}{c}{Zero-shot} & & \multicolumn{4}{c}{Few-shot}\\ 
            \cline{3-6}
            \cline{8-11}
             &  & EM & F1 & Entail. & Norm Entail. & & EM & F1 & Entail. & Norm Entail.\\
            \midrule
            \multirow{2}{*}{Alpaca-7B} & Open-Book & 41.64 & 51.97 & 74.47 & 79.74 && 31.83 & 43.41 & 68.67 & 73.85\\
            & Faithful Prompt & 46.05 & 56.60 & 76.28 & 79.74 && 49.55 & 61.97 & 78.08 & 79.62\\
            \midrule
            \multirow{2}{*}{ChatGPT} & Open-Book & 60.06 & 71.97 & 84.38 & 84.38 && 55.96 & 68.57 & 81.08 & 81.08\\
            & Faithful Prompt & 55.16 & 66.97 & 80.38 & 80.38 && 56.86 & 68.95 & 81.18 & 81.18\\
            \midrule
            \multirow{2}{*}{Palm} & Open-Book & 56.26 & 65.46 & 73.47 & 75.03 && 67.07 & 74.35 & 78.78 & 80.10\\
            & Faithful Prompt & 72.17 & 79.54 & 82.78 & 84.46 && 72.97 & 79.18 & 83.18 & 84.87\\
            \midrule
            \multirow{2}{*}{GPT-4} & Open-Book & 66.97 & 77.81 & 88.59 & 90.88 && 66.17 & 77.90 & 88.39 & 90.04\\
            & Faithful Prompt & 66.07 & 76.57 & 86.89 & 89.31 && 70.77 & 80.79 & 88.99 & 91.19\\
           
            \bottomrule
        \end{tabular}
        \caption{Zero-shot and few-shot performance of LLMs on Category 1 data of RealtimeQA.}
        \label{tab:3.3_cat1_result_realtime}
    \end{table*}

\section{Experiments}
\label{sec:experiment}

\subsection{Experiment Settings}
\label{ssec:setting}

\paragraph{Evaluation Metrics.} 
Three evaluation metrics are reported, \ie exact match (EM) accuracy, token-level F1, and entailment accuracy.
The first two metrics are traditionally used for evaluating QA models.
However, they tend to be too strict for evaluating LLM-generated responses, 
since LLMs often produce long and verbose sequences to explain the answers
(partially due to their alignment procedure).
The entailment accuracy is a more lenient metric that checks whether ``Question + LLM Output'' can entail ``Question + Answer''.
In this paper, we use an entailment model \texttt{nli-deberta-v3-base}\footnote{\url{https://huggingface.co/cross-encoder/nli-deberta-v3-base}} 
from Sentence-BERT \citep{reimers-2019-sentence-bert}, which is mostly reliable based on our manual inspection. Since we use the pivot model to select the seed cases, the accuracy of other models on the original set is not guaranteed to be 100. To clearly reveal the performance difference, we also report ``Normalized Entailment'' accuracy, where we normalize the test set to the cases that the corresponding model could answer correctly before perturbation. 

\paragraph{Source Data.} 
We use the MRQA version \citep{fisch2019mrqa} of Natural Questions  \citep{kwiatkowski-etal-2019-natural} and RealTimeQA data \citep{kasai2022realtime} from 20220613 to 20231110.
and conduct the following filtering steps:
1) remove duplicated Question-Evidence-Answer triplets and only keep one unique instance,
2) remove all evidence passages that are shorter than 10 words,
3) remove all cases with answers longer than 5 words.
After this, 7189 instances from NQ and 1380 instances from RealtimeQA are kept.
For questions with multiple answers, if the answers are overlapping (\eg ``1871'' and ``1871 A.D.''), we randomly keep one,
otherwise, the corresponding examples are removed.
Note the same question may still appear in multiple instances because the supporting evidence can be different.

\paragraph{Generated Data.}
Unless otherwise specified, ChatGPT (\texttt{gpt-3.5-turbo-0301}) is the pivot LLM for identifying seed test cases and generating attacking test cases.
When identifying seed test cases, we treat an answer produced by the pivot LLM as correct if 
it matches the reference answer exactly or can entail the reference answer in the same way as we compute the entailment accuracy.
The retriever used for generating Category 2 cases is based on \texttt{all-mpnet-base-v2}\footnote{\url{https://huggingface.co/sentence-transformers/all-mpnet-base-v2}}.
In total, we obtain \textbf{3,539} and \textbf{2,211} attacking test cases in Category 1 and Category 2 of NQ, and \textbf{1,000} and \textbf{814} attacking test cases in Category 1 and Category 2 of RealtimeQA respectively.

We evaluate five popular LLMs using the generated attacking test cases: Alpaca-7B \citep{alpaca}, ChatGPT (\texttt{gpt-3.5-turbo-0301}), Claude2, PaLM, and GPT-4 (\texttt{gpt-4-0613}), which is considered to be the state-of-the-art (SOTA) LLM.
In the few-shot setting, 5 static demonstration examples are used.

\subsection{Main Results}
\label{ssec:main_results}

\begin{table*}[ht]
        \small
        \setlength{\tabcolsep}{5pt}
        \centering
        \begin{tabular}{llcccccccccc}
            \toprule
            \multirow{2}{*}{Models} & \multirow{2}{*}{Method} & \multicolumn{4}{c}{Zero-shot} & & \multicolumn{4}{c}{Few-shot} \\ 
            \cline{3-6}
            \cline{8-11}
            
             &  & EM & F1 & Entail. & Norm Entail. & & EM & F1 & Entail. & Norm Entail.\\
            \midrule
            \multirow{2}{*}{Alpaca-7B} & Open-Book & 9.27 & 39.35 & 42.79 & 56.48 && 14.52 & 45.56 & 47.40 & 58.53\\
            & Faithful Prompt & 15.06 & 43.65 & 42.65 & 54.10 && 20.58 & 53.40 & 50.88 & 60.41\\
            \midrule
            \multirow{2}{*}{ChatGPT} & Open-Book & 25.51 & 57.15 & 61.78 & 61.78 && 27.32 & 58.94 & 51.15 & 51.15\\
            & Faithful Prompt & 24.69 & 53.49 & 50.38 & 50.38 && 24.20 & 56.26 & 44.10 & 44.10\\
            \midrule
            \multirow{2}{*}{Claude 2} & Open-Book & 29.99 & 58.69 & 43.46 & - && 12.12 & 39.83 & 57.26 & -\\
            & Faithful Prompt & 35.78 & 64.89 & 52.60 & - && 27.45 & 54.31 & 54.68 & -\\
            \midrule
            \multirow{2}{*}{Palm} & Open-Book & 44.78 & 71.76 & 66.76 & 75.70 && 50.84 & 75.23 & 66.53 & 75.58\\
            & Faithful Prompt & 44.78 & 70.18 & 58.75 & 66.03 && 47.35 & 72.03 & 61.78 & 69.01\\
            \midrule
            \multirow{2}{*}{GPT-4} & Open-Book & 37.68 & 67.27 & 68.39 & 73.55 && 46.27 & 74.17 & 73.04 & 77.95\\
            & Faithful Prompt & 33.60 & 62.78 & 58.25 & 62.36 && 45.59 & 72.83 & 67.57 & 72.46\\
           
            \bottomrule
        \end{tabular}
        \caption{Zero-shot and few-shot performance of LLMs on Category 2 Data of NQ.} 
        \label{tab:3.3_cat2_result}
    \end{table*}

\begin{table*}[ht]
        \small
        \setlength{\tabcolsep}{5pt}
        \centering
        \begin{tabular}{llcccccccccc}
            \toprule
            \multirow{2}{*}{Models} & \multirow{2}{*}{Method} & \multicolumn{4}{c}{Zero-shot} & & \multicolumn{4}{c}{Few-shot} \\ 
            \cline{3-6}
            \cline{8-11}
            
             &  & EM & F1 & Entail. & Norm Entail. & & EM & F1 & Entail. & Norm Entail.\\
            \midrule
            \multirow{2}{*}{Alpaca-7B} & Open-Book & 31.57 & 54.69 & 71.50 & 77.87 && 25.92 & 46.43 & 49.02 & 52.69\\
            & Faithful Prompt & 38.45 & 59.31 & 71.50 & 75.55 && 25.80 & 49.18 & 60.44 & 63.03\\
            \midrule
            \multirow{2}{*}{ChatGPT} & Open-Book & 42.87 & 63.77 & 72.73 & 72.73 && 36.00 & 56.56 & 64.25 & 64.25\\
            & Faithful Prompt & 31.82 & 53.06 & 60.81 & 60.81 && 44.10 & 66.63 & 75.18 & 75.18 \\
            \midrule
            \multirow{2}{*}{Palm} & Open-Book & 62.04 & 79.14 & 89.31 & 91.43 && 59.46 & 79.09 & 85.38 & 87.47\\
            & Faithful Prompt & 67.20 & 82.51 & 85.87 & 87.72 && 64.86 & 82.09 & 84.15 & 86.06\\
            \midrule
            \multirow{2}{*}{GPT-4} & Open-Book & 50.37 & 71.45 & 78.38 & 80.18 && 58.48 & 77.26 & 86.12 & 87.13\\
            & Faithful Prompt & 41.65 & 61.53 & 65.36 & 66.92 && 57.49 & 76.28 & 82.43 & 84.04\\
           
            \bottomrule
        \end{tabular}
        \caption{Zero-shot and few-shot performance of LLMs on Category 2 Data of RealtimeQA.} 
        \label{tab:3.3_cat2_result_realtime}
    \end{table*}

We evaluate the five LLMs on the Category 1 and Category 2 data generated by ChatGPT, including both self-attack and cross-attack scenarios. \footnote{Some numbers of Claude 2 are missing because we lost the access to the model due to Anthropic policy.}
In addition to vanilla zero-shot and few-shot promptings, we consider the recently proposed faithfulness-promoting prompting, \ie the opinion-based prompt by \cite{zhou2023context}.
For each model, we evaluate its closed-book performance, open-book performance, and open-book with faithful prompting performance. The full list of various prompts and error examples is in Appendix.

\paragraph{Category 1.} Here, the model is expected to follow the given context, and predict the \textit{altenative answer} proposed by the pivot model.
The results are summarized in \autoref{tab:3.3_cat1_result} and \autoref{tab:3.3_cat1_result_realtime}.
As expected, the model resistance towards our attack is mostly correlated with its model size and capability.
Specifically, larger and more capable models are more robust, \eg GPT-4 is more reliable than Alpaca-7B. 
Although GPT-4 is the most powerful model, it is still not immune to our attacks, indicating the effectiveness of our approach to trigger hallucination in SOTA LLMs.
Though using the human-designed faithful prompt or using in-context examples helps the performance in some cases, there are no consistent improvements compared with zero-shot in general.

\paragraph{Category 2.}
We require the model to fully understand both the question-focused expansion and evidence-focused expansion cases, and one question is considered correct only when both are answered correctly.
We report the merged result in Table \ref{tab:3.3_cat2_result} and Table \ref{tab:3.3_cat2_result_realtime}, and we also report the few-shot performance on each case separately in Table \ref{tab:3.3_cat2_single_result} of Appendix. 
As we can see, there are large performance drops for all models, suggesting they fail to identify the relevant evidence information regardless of prompting techniques.
Similar to Category 1, the faithful prompt is observed to have no consistent benefits, which calls for future work to develop more reliable prompting techniques.

\begin{table*}[ht]
        \small
        \setlength{\tabcolsep}{5pt}
        \centering
        \begin{tabular}{llccccccccccc}
            \toprule
            \multirow{2}{*}{Models} & \multirow{2}{*}{Method} & \multicolumn{3}{c}{ChatGPT} & & \multicolumn{3}{c}{GPT-4} & & \multicolumn{3}{c}{Alpaca-7B} \\ 
            \cline{3-5}
            \cline{7-9}
            \cline{11-13}
             &  & EM & F1 & Entail. & & EM & F1 & Entail. & & EM & F1 & Entail. \\
            \midrule
            \multirow{2}{*}{Alpaca-7B} & Open-Book & 25.00 & 40.57 & 61.20 && 26.8 & 43.88 & 68.2 && 26.00 & 43.95 & 65.80\\
            & Faithful Prompt & 37.20 & 53.46 & 72.20 && 39.60 & 57.49 & 76.00 && 36.60 & 53.93 & 70.80\\
            \midrule
            \multirow{2}{*}{ChatGPT} & Open-Book & 43.00 & 54.88 & 66.20 && 49.60 & 61.55 & 71.60 && 38.40 & 51.56 & 61.40\\
            & Faithful Prompt & 42.80 & 53.25 & 61.80 && 51.40 & 61.53 & 70.40 && 40.00 & 52.57 & 61.20\\
            \midrule
            \multirow{2}{*}{Palm} & Open-Book & 70.80 & 78.51 & 81.40 && 75.80 & 82.58 & 86.00 && 67.00 & 74.55 & 79.00\\
            & Faithful Prompt & 74.20 & 82.00 & 84.40 &&  78.80 & 85.28 & 89.00 && 69.20 & 77.73 & 82.80\\
            \midrule
            \multirow{2}{*}{GPT-4} & Open-Book & 65.20 & 76.66 & 84.00 && 59.20 & 69.18 & 76.40 && 57.00 & 67.23 & 73.80\\
            & Faithful Prompt & 69.80 & 79.04 & 84.80 && 67.40 & 75.98 & 81.80 && 59.60 & 70.15 & 78.40\\
           
            \bottomrule
        \end{tabular}
        \caption{Few-shot case study of backbone LLMs used by \ourmodel (500 examples). The column blocks indicate the Category 1 data generated by ChatGPT, GPT-4, and Alpaca-7B, respectively} 
        \label{tab:3.4_cat1_result}
    \end{table*}

\begin{table*}[ht]
        \small
        \setlength{\tabcolsep}{5pt}
        \centering
        \begin{tabular}{llcccccccc}
            \toprule
            \multirow{2}{*}{Models} & \multirow{2}{*}{Method} & \multicolumn{3}{c}{ChatGPT} & & \multicolumn{3}{c}{GPT-4} \\ 
            \cline{3-5}
            \cline{7-9}
            
             &  & EM & F1 & Entail. & & EM & F1 & Entail. \\
            \midrule
            \multirow{2}{*}{Alpaca-7B} & Open-Book & 17.80 & 44.85 & 52.20 && 9.00 & 37.16 & 42.40\\
            & Faithful Prompt & 22.40 & 53.96 & 57.00 && 16.00 & 46.28 & 43.80\\
            \midrule
            \multirow{2}{*}{ChatGPT} & Open-Book & 29.40 & 57.12 & 50.80 && 23.20 & 50.76 & 46.20\\
            & Faithful Prompt & 24.40 & 54.61 & 41.60 && 23.20 & 52.80 & 43.20\\
            \midrule
            \multirow{2}{*}{Palm} & Open-Book & 54.40 & 76.84 & 69.60 && 52.20 & 73.62 & 66.40\\
            & Faithful Prompt & 53.40 & 75.93 & 68.60 && 48.4 & 71.91 & 62.60\\
            \midrule
            \multirow{2}{*}{GPT-4} & Open-Book & 49.40 & 74.38 & 74.20 && 24.00 & 47.18 & 37.60\\
            & Faithful Prompt & 51.80 & 73.68 & 71.00 && 35.00 & 62.04 & 52.40\\
           
            \bottomrule
        \end{tabular}
        \caption{Few-shot case study of backbone LLMs used by \ourmodel (500 examples). The column blocks indicate
the Category 2 data generated by ChatGPT and GPT-4, respectively}
        \label{tab:3.4_cat2_multi_result}
    \end{table*}

\subsection{Human Evaluations}
\label{sec:mturk}

To evaluate whether the evidence generated by \ourmodel is supportive and human-readable,
we randomly sample 500 cases from Category 1, 1000 cases from Category 2 with 500 examples for question-focused expansion, and 500 for evidence-focused expansion.
We use Amazon Mechanical Turk to collect human judgments on this set.
Each question is judged by three annotators, who are asked to read the evidence and decide whether it could support them to get the correct answer.
To prevent annotators from randomly submitting ``Yes'' or ``No'', 10\% of the data is used as validation checks where we know whether the evidence supports the answer.
We only accept annotations from the annotators with at least 90\% accuracy on the validation check.
For each question, if the majority of the annotators think the generated evidence is supportive, it is then counted as human-readable.
For all three categories, around 90\% of the cases are human readable, supporting the quality of \ourmodel,
with $90.8\%$, $92.4\%$, and $88.8\%$ human-readable ratios for Category 1, Category 2 question-focused and evidence-focused, respectively.

\subsection{Case Studies}
\paragraph{Is \ourmodel sensitive toward backbone LLMs?}
\label{sec:backbone}

To do that, we use alternative LLMs to generate attacking test cases other than ChatGPT.
We consider both Alpaca-7b and GPT-4 for Category 1 and only GPT-4 for Category 2 given the task is more demanding.
Due to the limitation of budget, we randomly sample 500 examples from NQ for this study. All prompts are similar to those used previously.
The few-shot performances of Category 1 and Category 2 are reported in \autoref{tab:3.4_cat1_result} and \autoref{tab:3.4_cat2_multi_result}, respectively.
As shown in \autoref{tab:3.4_cat1_result}, compared with ChatGPT and Alpaca, GPT-4 does not generate stronger attacks.
This is probably because the alternative answers from GPT-4 are more receptive to all models.
The Category 1 data generated by the smallest model (Alpaca-7B) appears to be very effective for those two larger ones, but we observe that that is because Alpaca sometimes generates invalid answers and also fails to replace all the occurrences of the old answer.
On the other hand, compared with ChatGPT, GPT-4 can generate more stronger attacks for Category 2 (\autoref{tab:3.4_cat2_multi_result}).
We find that GPT-4 is better at summarizing multiple pieces of information, leading to more complex evidence.
Although all three models are most vulnerable to self-attacks, all \ourmodel attacks are transferable, making it possible to generate attacking test cases using more cost-effective models.

\paragraph{Is \ourmodel sensitive toward the position of the answer?}
\label{sec:position_bias}

To get the distribution of the answer in the evidence, we only keep the cases where the answer only occurs once in the evidence (2678 in total). There are $55.94\%$ cases where the answer is in the first 1/3 of the evidence, $23.64\%$ cases where the answer is in the middle part of the evidence, and $20.43\%$ cases where the answer is in the last 1/3 of the evidence. We evaluate the accuracy of different models under both few-shot and open-book setting in these 3 cases, and we do not see any significant performance difference except that Alpaca-7B performs worse when the answer is at the end of the evidence. More detailed results are in \autoref{tab:4.4_position_result} in the Appendix.

\section{Conclusion}
In this paper, we present \ourmodel, an LLM-based framework that generates transferable adversarial attacks to assess the hallucination of retrieval-augmented LLMs.
By swapping the answer in the evidence or adding more relevant information to enrich the context, we successfully trigger hallucination behaviors of existing state-of-the-art LLMs.
\ourmodel is a viable approach in that it can generate transferable attacking examples using more cost-effective LLMs.
We believe \ourmodel could be used to help assess the hallucination of future LLMs, and potentially help mitigate hallucinations.  
Future directions include further studying \ourmodel on tasks of different complexities and how to use \ourmodel for debugging LLM-based applications.

\section{Limitations}
Although we find our framework effective in evaluating the reliability of retrieval-augmented LLMs, there are some limitations worth discussion here.

First, this study distinctly concentrates on questions with short answers, thereby delineating an intentional boundary from engaging in the exploration of long-form question-answering. For long-form cases, it requires more complex ways of perturbing evidence, \eg multiple sentences are required to be updated at the same time. The comprehensive investigation into long-form question-answering is deferred to future scholarly endeavors, marking a deliberate scope restriction to refine the focus and depth of the current analysis. 

Moreover, the scope of our research rigorously limits its examination to single-hop questions. Consequently, this study does not venture into the evaluation of complex reasoning inaccuracies, often referred to as reasoning hallucinations, which is more likely for multi-hop questions. This delineation underscores a focused approach, yet acknowledges the complexity and necessity of future investigations into multi-hop question-answering, with the need for specialized methodologies to assess and mitigate reasoning errors in such contexts.

In terms of methodology, our study either introduces perturbations within the answer span or modifies the adjacent contextual narrative; however, scenarios that encompass both an altered answer span and a significantly adjusted surrounding context are not within the purview of this investigation. This strategic decision enables the isolation and better understanding of the effects of each type of perturbation independently. Nonetheless, it also marks a critical avenue for further intricate research toward evaluating the compound impacts.

\bibliography{custom}

\begin{thebibliography}{30}
\expandafter\ifx\csname natexlab\endcsname\relax\def\natexlab#1{#1}\fi

\bibitem[{Anil et~al.(2023)Anil, Dai, Firat, Johnson, Lepikhin, Passos, Shakeri, Taropa, Bailey, Chen, Chu, Clark, Shafey, Huang, Meier-Hellstern, Mishra, Moreira, Omernick, Robinson, Ruder, Tay, Xiao, Xu, Zhang, Abrego, Ahn, Austin, Barham, Botha, Bradbury, Brahma, Brooks, Catasta, Cheng, Cherry, Choquette-Choo, Chowdhery, Crepy, Dave, Dehghani, Dev, Devlin, Díaz, Du, Dyer, Feinberg, Feng, Fienber, Freitag, Garcia, Gehrmann, Gonzalez, Gur-Ari, Hand, Hashemi, Hou, Howland, Hu, Hui, Hurwitz, Isard, Ittycheriah, Jagielski, Jia, Kenealy, Krikun, Kudugunta, Lan, Lee, Lee, Li, Li, Li, Li, Li, Lim, Lin, Liu, Liu, Maggioni, Mahendru, Maynez, Misra, Moussalem, Nado, Nham, Ni, Nystrom, Parrish, Pellat, Polacek, Polozov, Pope, Qiao, Reif, Richter, Riley, Ros, Roy, Saeta, Samuel, Shelby, Slone, Smilkov, So, Sohn, Tokumine, Valter, Vasudevan, Vodrahalli, Wang, Wang, Wang, Wang, Wieting, Wu, Xu, Xu, Xue, Yin, Yu, Zhang, Zheng, Zheng, Zhou, Zhou, Petrov, and Wu}]{anil2023palm}
Rohan Anil, Andrew~M. Dai, Orhan Firat, Melvin Johnson, Dmitry Lepikhin, Alexandre Passos, Siamak Shakeri, Emanuel Taropa, Paige Bailey, Zhifeng Chen, Eric Chu, Jonathan~H. Clark, Laurent~El Shafey, Yanping Huang, Kathy Meier-Hellstern, Gaurav Mishra, Erica Moreira, Mark Omernick, Kevin Robinson, Sebastian Ruder, Yi~Tay, Kefan Xiao, Yuanzhong Xu, Yujing Zhang, Gustavo~Hernandez Abrego, Junwhan Ahn, Jacob Austin, Paul Barham, Jan Botha, James Bradbury, Siddhartha Brahma, Kevin Brooks, Michele Catasta, Yong Cheng, Colin Cherry, Christopher~A. Choquette-Choo, Aakanksha Chowdhery, Clément Crepy, Shachi Dave, Mostafa Dehghani, Sunipa Dev, Jacob Devlin, Mark Díaz, Nan Du, Ethan Dyer, Vlad Feinberg, Fangxiaoyu Feng, Vlad Fienber, Markus Freitag, Xavier Garcia, Sebastian Gehrmann, Lucas Gonzalez, Guy Gur-Ari, Steven Hand, Hadi Hashemi, Le~Hou, Joshua Howland, Andrea Hu, Jeffrey Hui, Jeremy Hurwitz, Michael Isard, Abe Ittycheriah, Matthew Jagielski, Wenhao Jia, Kathleen Kenealy, Maxim Krikun, Sneha Kudugunta, Chang
  Lan, Katherine Lee, Benjamin Lee, Eric Li, Music Li, Wei Li, YaGuang Li, Jian Li, Hyeontaek Lim, Hanzhao Lin, Zhongtao Liu, Frederick Liu, Marcello Maggioni, Aroma Mahendru, Joshua Maynez, Vedant Misra, Maysam Moussalem, Zachary Nado, John Nham, Eric Ni, Andrew Nystrom, Alicia Parrish, Marie Pellat, Martin Polacek, Alex Polozov, Reiner Pope, Siyuan Qiao, Emily Reif, Bryan Richter, Parker Riley, Alex~Castro Ros, Aurko Roy, Brennan Saeta, Rajkumar Samuel, Renee Shelby, Ambrose Slone, Daniel Smilkov, David~R. So, Daniel Sohn, Simon Tokumine, Dasha Valter, Vijay Vasudevan, Kiran Vodrahalli, Xuezhi Wang, Pidong Wang, Zirui Wang, Tao Wang, John Wieting, Yuhuai Wu, Kelvin Xu, Yunhan Xu, Linting Xue, Pengcheng Yin, Jiahui Yu, Qiao Zhang, Steven Zheng, Ce~Zheng, Weikang Zhou, Denny Zhou, Slav Petrov, and Yonghui Wu. 2023.
\newblock \href {http://arxiv.org/abs/2305.10403} {Palm 2 technical report}.

\bibitem[{Anthropic(2023)}]{claude2}
Anthropic. 2023.
\newblock \href {https://www.anthropic.com/index/claude-2} {Claude 2}.

\bibitem[{Carlini et~al.(2023)Carlini, Nasr, Choquette-Choo, Jagielski, Gao, Awadalla, Koh, Ippolito, Lee, Tramer, and Schmidt}]{carlini2023aligned}
Nicholas Carlini, Milad Nasr, Christopher~A. Choquette-Choo, Matthew Jagielski, Irena Gao, Anas Awadalla, Pang~Wei Koh, Daphne Ippolito, Katherine Lee, Florian Tramer, and Ludwig Schmidt. 2023.
\newblock \href {http://arxiv.org/abs/2306.15447} {Are aligned neural networks adversarially aligned?}

\bibitem[{Cheng et~al.(2021)Cheng, Liu, Pereira, Yu, and Gao}]{cheng-etal-2021-posterior}
Hao Cheng, Xiaodong Liu, Lis Pereira, Yaoliang Yu, and Jianfeng Gao. 2021.
\newblock \href {https://doi.org/10.18653/v1/2021.naacl-main.85} {Posterior differential regularization with f-divergence for improving model robustness}.
\newblock In \emph{Proceedings of the 2021 Conference of the North American Chapter of the Association for Computational Linguistics: Human Language Technologies}, pages 1078--1089, Online. Association for Computational Linguistics.

\bibitem[{Choi et~al.(2021)Choi, Palomaki, Lamm, Kwiatkowski, Das, and Collins}]{choi2021decontextualization}
Eunsol Choi, Jennimaria Palomaki, Matthew Lamm, Tom Kwiatkowski, Dipanjan Das, and Michael Collins. 2021.
\newblock Decontextualization: Making sentences stand-alone.
\newblock \emph{Transactions of the Association for Computational Linguistics}, 9:447--461.

\bibitem[{Fisch et~al.(2019)Fisch, Talmor, Jia, Seo, Choi, and Chen}]{fisch2019mrqa}
Adam Fisch, Alon Talmor, Robin Jia, Minjoon Seo, Eunsol Choi, and Danqi Chen. 2019.
\newblock {MRQA} 2019 shared task: Evaluating generalization in reading comprehension.
\newblock In \emph{Proceedings of 2nd Machine Reading for Reading Comprehension (MRQA) Workshop at EMNLP}.

\bibitem[{Goodfellow et~al.(2014)Goodfellow, Shlens, and Szegedy}]{goodfellow2014explaining}
Ian~J Goodfellow, Jonathon Shlens, and Christian Szegedy. 2014.
\newblock Explaining and harnessing adversarial examples.
\newblock \emph{arXiv preprint arXiv:1412.6572}.

\bibitem[{Iyyer et~al.(2018)Iyyer, Wieting, Gimpel, and Zettlemoyer}]{iyyer-etal-2018-adversarial}
Mohit Iyyer, John Wieting, Kevin Gimpel, and Luke Zettlemoyer. 2018.
\newblock \href {https://doi.org/10.18653/v1/N18-1170} {Adversarial example generation with syntactically controlled paraphrase networks}.
\newblock In \emph{Proceedings of the 2018 Conference of the North {A}merican Chapter of the Association for Computational Linguistics: Human Language Technologies, Volume 1 (Long Papers)}, pages 1875--1885, New Orleans, Louisiana. Association for Computational Linguistics.

\bibitem[{Jia and Liang(2017)}]{jia-liang-2017-adversarial}
Robin Jia and Percy Liang. 2017.
\newblock \href {https://doi.org/10.18653/v1/D17-1215} {Adversarial examples for evaluating reading comprehension systems}.
\newblock In \emph{Proceedings of the 2017 Conference on Empirical Methods in Natural Language Processing}, pages 2021--2031, Copenhagen, Denmark. Association for Computational Linguistics.

\bibitem[{Kasai et~al.(2022)Kasai, Sakaguchi, Takahashi, Bras, Asai, Yu, Radev, Smith, Choi, and Inui}]{kasai2022realtime}
Jungo Kasai, Keisuke Sakaguchi, Yoichi Takahashi, Ronan~Le Bras, Akari Asai, Xinyan Yu, Dragomir Radev, Noah~A Smith, Yejin Choi, and Kentaro Inui. 2022.
\newblock Realtime qa: What's the answer right now?
\newblock \emph{arXiv preprint arXiv:2207.13332}.

\bibitem[{Kwiatkowski et~al.(2019)Kwiatkowski, Palomaki, Redfield, Collins, Parikh, Alberti, Epstein, Polosukhin, Devlin, Lee, Toutanova, Jones, Kelcey, Chang, Dai, Uszkoreit, Le, and Petrov}]{kwiatkowski-etal-2019-natural}
Tom Kwiatkowski, Jennimaria Palomaki, Olivia Redfield, Michael Collins, Ankur Parikh, Chris Alberti, Danielle Epstein, Illia Polosukhin, Jacob Devlin, Kenton Lee, Kristina Toutanova, Llion Jones, Matthew Kelcey, Ming-Wei Chang, Andrew~M. Dai, Jakob Uszkoreit, Quoc Le, and Slav Petrov. 2019.
\newblock \href {https://doi.org/10.1162/tacl_a_00276} {Natural questions: A benchmark for question answering research}.
\newblock \emph{Transactions of the Association for Computational Linguistics}, 7:452--466.

\bibitem[{Li and Qiu(2020)}]{li2020textat}
Linyang Li and Xipeng Qiu. 2020.
\newblock Textat: Adversarial training for natural language understanding with token-level perturbation.
\newblock \emph{arXiv preprint arXiv:2004.14543}.

\bibitem[{Longpre et~al.(2021)Longpre, Perisetla, Chen, Ramesh, DuBois, and Singh}]{longpre2021entity}
Shayne Longpre, Kartik Perisetla, Anthony Chen, Nikhil Ramesh, Chris DuBois, and Sameer Singh. 2021.
\newblock Entity-based knowledge conflicts in question answering.
\newblock \emph{arXiv preprint arXiv:2109.05052}.

\bibitem[{Madry et~al.(2018)Madry, Makelov, Schmidt, Tsipras, and Vladu}]{madry2017pgd}
Aleksander Madry, Aleksandar Makelov, Ludwig Schmidt, Dimitris Tsipras, and Adrian Vladu. 2018.
\newblock \href {https://openreview.net/forum?id=rJzIBfZAb} {Towards deep learning models resistant to adversarial attacks}.
\newblock In \emph{International Conference on Learning Representations}.

\bibitem[{Nie et~al.(2020)Nie, Williams, Dinan, Bansal, Weston, and Kiela}]{nie-etal-2020-adversarial}
Yixin Nie, Adina Williams, Emily Dinan, Mohit Bansal, Jason Weston, and Douwe Kiela. 2020.
\newblock \href {https://doi.org/10.18653/v1/2020.acl-main.441} {Adversarial {NLI}: A new benchmark for natural language understanding}.
\newblock In \emph{Proceedings of the 58th Annual Meeting of the Association for Computational Linguistics}, pages 4885--4901, Online. Association for Computational Linguistics.

\bibitem[{OpenAI(2022)}]{openai2022chatgpt}
OpenAI. 2022.
\newblock \href {https://openai.com/blog/chatgpt} {{C}hat{GPT}}.

\bibitem[{OpenAI(2023)}]{openai2023gpt4}
OpenAI. 2023.
\newblock {GPT-4} technical report.
\newblock \emph{arXiv preprint arXiv:2303.08774}.

\bibitem[{Papernot et~al.(2016)Papernot, McDaniel, and Goodfellow}]{papernot2016transferability}
Nicolas Papernot, Patrick McDaniel, and Ian Goodfellow. 2016.
\newblock \href {http://arxiv.org/abs/1605.07277} {Transferability in machine learning: from phenomena to black-box attacks using adversarial samples}.

\bibitem[{Peng et~al.(2023)Peng, Galley, He, Cheng, Xie, Hu, Huang, Liden, Yu, Chen, and Gao}]{peng2023check}
Baolin Peng, Michel Galley, Pengcheng He, Hao Cheng, Yujia Xie, Yu~Hu, Qiuyuan Huang, Lars Liden, Zhou Yu, Weizhu Chen, and Jianfeng Gao. 2023.
\newblock \href {http://arxiv.org/abs/2302.12813} {Check your facts and try again: Improving large language models with external knowledge and automated feedback}.

\bibitem[{Reimers and Gurevych(2019)}]{reimers-2019-sentence-bert}
Nils Reimers and Iryna Gurevych. 2019.
\newblock \href {http://arxiv.org/abs/1908.10084} {Sentence-bert: Sentence embeddings using siamese bert-networks}.
\newblock In \emph{Proceedings of the 2019 Conference on Empirical Methods in Natural Language Processing}. Association for Computational Linguistics.

\bibitem[{Shi et~al.(2023)Shi, Min, Yasunaga, Seo, James, Lewis, Zettlemoyer, and tau Yih}]{shi2023replug}
Weijia Shi, Sewon Min, Michihiro Yasunaga, Minjoon Seo, Rich James, Mike Lewis, Luke Zettlemoyer, and Wen tau Yih. 2023.
\newblock \href {http://arxiv.org/abs/2301.12652} {Replug: Retrieval-augmented black-box language models}.

\bibitem[{Szegedy et~al.(2014)Szegedy, Zaremba, Sutskever, Bruna, Erhan, Goodfellow, and Fergus}]{szegedy2014intriguing}
Christian Szegedy, Wojciech Zaremba, Ilya Sutskever, Joan Bruna, Dumitru Erhan, Ian Goodfellow, and Rob Fergus. 2014.
\newblock Intriguing properties of neural networks.
\newblock In \emph{International Conference on Learning Representations}.

\bibitem[{Taori et~al.(2023)Taori, Gulrajani, Zhang, Dubois, Li, Guestrin, Liang, and Hashimoto}]{alpaca}
Rohan Taori, Ishaan Gulrajani, Tianyi Zhang, Yann Dubois, Xuechen Li, Carlos Guestrin, Percy Liang, and Tatsunori~B. Hashimoto. 2023.
\newblock Stanford alpaca: An instruction-following llama model.
\newblock \url{https://github.com/tatsu-lab/stanford_alpaca}.

\bibitem[{Tsipras et~al.(2019)Tsipras, Santurkar, Engstrom, Turner, and Madry}]{tsipras2018robustness}
Dimitris Tsipras, Shibani Santurkar, Logan Engstrom, Alexander Turner, and Aleksander Madry. 2019.
\newblock \href {https://openreview.net/forum?id=SyxAb30cY7} {Robustness may be at odds with accuracy}.
\newblock In \emph{International Conference on Learning Representations}.

\bibitem[{Wallace et~al.(2021)Wallace, Feng, Kandpal, Gardner, and Singh}]{wallace2021universal}
Eric Wallace, Shi Feng, Nikhil Kandpal, Matt Gardner, and Sameer Singh. 2021.
\newblock \href {http://arxiv.org/abs/1908.07125} {Universal adversarial triggers for attacking and analyzing nlp}.

\bibitem[{Xie et~al.(2023)Xie, Zhang, Chen, Lou, and Su}]{xie2023adaptive}
Jian Xie, Kai Zhang, Jiangjie Chen, Renze Lou, and Yu~Su. 2023.
\newblock Adaptive chameleon or stubborn sloth: Unraveling the behavior of large language models in knowledge conflicts.
\newblock \emph{arXiv preprint arXiv:2305.13300}.

\bibitem[{Yan et~al.(2021)Yan, Xiao, Mukherjee, Lin, Jia, and Ren}]{yan2021robustness}
Jun Yan, Yang Xiao, Sagnik Mukherjee, Bill~Yuchen Lin, Robin Jia, and Xiang Ren. 2021.
\newblock On the robustness of reading comprehension models to entity renaming.
\newblock \emph{arXiv preprint arXiv:2110.08555}.

\bibitem[{Zhang et~al.(2019)Zhang, Yu, Jiao, Xing, Ghaoui, and Jordan}]{zhang2019theoretically}
Hongyang Zhang, Yaodong Yu, Jiantao Jiao, Eric Xing, Laurent~El Ghaoui, and Michael Jordan. 2019.
\newblock \href {http://proceedings.mlr.press/v97/zhang19p.html} {Theoretically principled trade-off between robustness and accuracy}.
\newblock In \emph{Proceedings of the 36th International Conference on Machine Learning}, volume~97 of \emph{Proceedings of Machine Learning Research}, pages 7472--7482. PMLR.

\bibitem[{Zhou et~al.(2023)Zhou, Zhang, Poon, and Chen}]{zhou2023context}
Wenxuan Zhou, Sheng Zhang, Hoifung Poon, and Muhao Chen. 2023.
\newblock Context-faithful prompting for large language models.
\newblock \emph{arXiv preprint arXiv:2303.11315}.

\bibitem[{Zhu et~al.(2020)Zhu, Cheng, Gan, Sun, Goldstein, and Liu}]{zhu2019freelb}
Chen Zhu, Yu~Cheng, Zhe Gan, Siqi Sun, Tom Goldstein, and Jingjing Liu. 2020.
\newblock \href {https://openreview.net/forum?id=BygzbyHFvB} {Freelb: Enhanced adversarial training for natural language understanding}.
\newblock In \emph{International Conference on Learning Representations}.

\end{thebibliography}

\appendix
\section{Appendix}
Here, we provide examples of prompt implementations and additional results.

First, we provide selected instances for few-shot prompting in \autoref{tab:few_shot_instances}.
In addition, prompts used for Category 1 and Category 2 data generations are listed in \autoref{tab:cat1_prompt} and \autoref{tab:cat2_prompt} respectively.
Different prompting methods for different language models are detailed in
\autoref{tab:closed-book_prompt} (Closed-book), \autoref{tab:open-book_prompt} (Open-book) and \autoref{tab:bob_prompt} (Faithful prompting).

Lastly, more experiment results are in \autoref{tab:3.3_cat2_single_result} (breakdown results of Category 2 experiments in \autoref{ssec:main_results}) and \autoref{tab:4.4_position_result} (detailed results for \autoref{sec:position_bias}).
\begin{table*}[ht!]
        \centering
        \begin{tabular}{l}
            \toprule
            Question: who sings what lovers do with maroon 5\\
            Evidence: `` What Lovers Do '' is a song by American pop rock band Maroon 5 featuring\\
            \hspace*{0.5cm} American R\&B singer SZA . It was released on August 30 , 2017 , as the lead single\\
            \hspace*{0.5cm} from the band 's sixth studio album Red Pill Blues ( 2017 ) . The song contains an\\
            \hspace*{0.5cm} interpolation of the 2016 song `` Sexual '' by Neiked featuring Dyo , therefore\\
            \hspace*{0.5cm} Victor Rådström , Dyo and Elina Stridh are credited as songwriters . \\
            Answer: American R\&B singer SZA\\
            \midrule
            Question: who plays lead guitar on i want you she 's so heavy\\
            Evidence: John Lennon -- lead and harmony vocals , multi-tracked lead guitar ,
            Moog \\
            \hspace*{0.5cm}synthesizer~~~~Paul McCartney -- harmony vocals, bass~~~~George Harrison -- harmony \\
            \hspace*{0.5cm}vocals , multi-tracked lead guitar~~~~Ringo Starr -- drums , congas , wind machine   Billy \\
            \hspace*{0.5cm}Preston -- Hammond organ\\
            Answer: John Lennon\\
            \midrule
            Question: a long chain of amino acids linked by peptide bonds is a\\
            Evidence: The covalent chemical bonds are formed when the carboxyl group of one amino \\
            \hspace*{0.5cm} acid reacts with the amino group of another . The shortest peptides are dipeptides ,\\
            \hspace*{0.5cm}  consisting of 2 amino acids joined by a single peptide bond , followed by tripeptides ,\\
            \hspace*{0.5cm}  tetrapeptides , etc . A polypeptide is a long , continuous , and unbranched peptide chain .\\
            \hspace*{0.5cm}  Hence , peptides fall under the broad chemical classes of biological oligomers and\\
            \hspace*{0.5cm}  polymers , alongside nucleic acids , oligosaccharides and polysaccharides , etc .\\
            Answer: polypeptide\\
            \midrule
            Question: when does the school year start in france\\
            Evidence: In Metropolitan France , the school year runs from early September to early July .\\
            \hspace*{0.5cm}  The school calendar is standardised throughout the country and is the sole domain of \\
            \hspace*{0.5cm} the ministry .\\
            Answer: early September\\
            \midrule
            Question: which city is selected under hriday scheme in karnataka\\
            Evidence: With a duration of 4 years ( completing in November 2018 ) and a total outlay of \\
            \hspace*{0.5cm}  500 crore ( US \$78 million ) , the Scheme is set to be implemented in 12 identified Cities\\
            \hspace*{0.5cm}  namely , Ajmer , Amaravati , Amritsar , Badami , Dwarka , Gaya , Kanchipuram , \\
            \hspace*{0.5cm}  Mathura , Puri , Varanasi , Velankanni and Warangal .\\
            Answer: Ajmer\\

            \bottomrule
        \end{tabular}
        \caption{Five Randomly Selected Demo Instances from NQ Training Data for Few-shot Experiments. } 
        \label{tab:few_shot_instances}
    \end{table*}

\begin{table*}[h!]
    \centering
    \resizebox{\textwidth}{!}{
    \begin{tabular}{l|l}
        \toprule
        & \\
        \multirow{12}{*}{Generate Alternative Answer Prompt}& A question and its correct answer is below. Generate\\
        &  a wrong answer to the question that is different from \\
        & the correct answer. Make sure the wrong answer is short, \\
        & and has the same type as the correct answer.\\
        & \\
        & Question: \\
        & \{Question\}\\
        & \\
        & Answer:\\
        & \{Answer\}\\
        & \\
        & Wrong Answer: \\
        \midrule 
        \multirow{14}{*}{Replace Old Answer Prompt } & A passage and a text span inside the passage is shown\\
        & below. Rewrite the passage to replace all the occurren-\\
        & ces of the text span with the new span.\\
        & \\
        & Passage: \\
        & \{Passage\}\\
        & \\
        & Text Span:\\
        & \{Answer\}\\
        & \\
        & New Span:\\
        & \{Alternative Answer\}\\
        & \\
        & New Passage: \\
        \bottomrule
    \end{tabular}}
    \caption{Prompts for Cat1 Data Generation.} 
    \label{tab:cat1_prompt}
\end{table*}

\begin{table*}[h!]
    \centering
    \begin{tabular}{l|l}
        \toprule
        & \\
        \multirow{14}{*}{Select Supporting Sentence Prompt} & A question, the answer, and a passage are shown below. \\
        & Please select the sentence in the passage that supports \\
        & to answer the question correctly.\\
        & \\
        & Question: \\
        & \{Question\}\\
        & \\
        & Answer: \\
        & \{Answer\} \\
        & \\
        & Passage: \\
        & \{Passage\} \\
        & \\
        & Sentence: \\
        \midrule
        \multirow{11}{*}{Summarize Relevant Passages Prompt} & Three relevant passages are shown below.\\
        &   Please condense the three passages into one passage.\\
        & \\
        & Relevant Passages: \\
        & [1]: \{Passage 1\}\\
        & \\
        & [2]: \{Passage 2\}\\
        & \\
        & [3]: \{Passage 3\}\\
        & \\
        & Relevant New Information: \\
        \midrule 
        \multirow{13}{*}{Merge Prompt } & Two passages and a span are shown below. Please \\
        & merge the two passages, and make sure to keep the\\
        & span in the new passage.\\
        & \\
        & Passages: \\
        & [1]: \{Supporting Sentence\}\\
        & \\
        & [2]: \{Summarized Passage\}\\
        & \\
        & Span:\\
        & \{Answer\}\\
        & \\
        & New Passage: \\
        \bottomrule
    \end{tabular}
    \caption{Prompts for Cat2 Data Generation.} 
    \label{tab:cat2_prompt}
\end{table*}

\begin{table*}[h!]
    \centering
    \begin{tabular}{l|l}
        \toprule
        & \\
        \multirow{8}{*}{Alpaca-7B} & Below is an instruction that describes a task.\\
        & Write a response that appropriately completes the request. \\
        & Only output the answer without other context words. \\
        & \\
        & \#\#\# Instruction:\\
        & \{Question\}\\
        & \\
        & \#\#\# Response: \\
        \midrule
        \multirow{11}{*}{PaLM} & You are a helpful and informative bot that answers questions \\
        & Be sure to respond in a complete sentence, being comprehensive, \\
        & including all relevant background information.   However, you\\
        & are talking to a non-technical audience, so be sure to break \\
        & down complicated concepts and   strike a friendly and convers-\\
        & tional tone.   Only output the answer without other context words.\\
        & \\
        & QUESTION:\\
        & \{Question\}\\
        & \\
        & ANSWER: \\
        \midrule 
        \multirow{8}{*}{Claude 2} & Human: \\
        & Answer the question below. Only output the answer without other\\
        &  context words.\\
        & \\
        & Question:\\
        & \{Question\}\\
        & \\
        & Assistant:\\
        \midrule
        \multirow{9}{*}{ChatGPT \& GPT-4} & system: You are a helpful assistant.\\
        & \\
        & user: Answer the question below. Only output the answer without other\\
        &  context words.\\
        & \\
        & Question:\\
        & \{Question\}\\
        & \\
        & Answer:\\
        \bottomrule
    \end{tabular}
    \caption{Closed-Book QA prompts for all considered models following their corresponding recommendations.} 
    \label{tab:closed-book_prompt}
\end{table*}

\begin{table*}[h!]
    \centering
    \small
    \begin{tabular}{l|l}
        \toprule
        & \\
        \multirow{12}{*}{Alpaca-7B} & Below is an instruction that describes a task, paired with\\
        & an input that provides further context.\\
        & Write a response that appropriately completes the request. \\
        & Only output the answer without other context words. \\
        & \\
        & \#\#\# Instruction:\\
        & \{Question\}\\
        & \\
        & \#\#\# Input:\\
        & \{Evidence\} \\
        & \\
        & \#\#\# Response: \\
        \midrule
        \multirow{16}{*}{PaLM} & You are a helpful and informative bot that answers questions \\
        & using text from the reference passage included below.   Be \\
        & sure to respond in a complete sentence, being comprehensive, \\
        & including all relevant background information.   However, you\\
        & are talking to a non-technical audience, so be sure to break \\
        & down complicated concepts and   strike a friendly and convers-\\
        & tional tone.   If the passage is irrelevant to the answer, you\\
        & may ignore it. Only output the answer without other context words.\\
        & \\
        & QUESTION:\\
        & \{Question\}\\
        & \\
        & PASSAGE:\\
        & \{Evidence\}\\
        & \\
        & ANSWER: \\
        \midrule 
        \multirow{12}{*}{Claude 2} & Human: \\
        & Answer the question below, paired with a context that provides\\
        &  background knowledge. Only output the answer without other\\
        &  context words.\\
        & \\
        & Context:\\
        & \{Evidence\}\\
        & \\
        & Question:\\
        & \{Question\}\\
        & \\
        & Assistant:\\
        \midrule
        \multirow{13}{*}{ChatGPT \& GPT-4} & system: You are a helpful assistant.\\
        & \\
        & user: Answer the question below, paired with a context that provides\\
        &  background knowledge. Only output the answer without other\\
        &  context words.\\
        & \\
        & Context:\\
        & \{Evidence\}\\
        & \\
        & Question:\\
        & \{Question\}\\
        & \\
        & Answer:\\
        \bottomrule
    \end{tabular}
    \caption{Open-Book Inference Prompts for Different Models Following their Official Instructions.} 
    \label{tab:open-book_prompt}
\end{table*}

\begin{table*}[h!]
   \centering
   \begin{tabular}{l|l}
       \toprule
       & \\
       \multirow{7}{*}{Alpaca-7B} & Instruction: read the given information and answer the corresponding \\
       & question. Only output the answer without other context words.\\
       & \\
       & \#\#\# Instruction: Bob said, ``\{Evidence\}''\\
       & Q: \{Question\} in Bob's opinion based on the given text?\\
       & \\
       & \#\#\# Response: \\
       \midrule
       \multirow{5}{*}{PaLM} & Instruction: read the given information and answer the corresponding \\
       & question. Only output the answer without other context words.\\
       & \\
       & Bob said, ``\{Evidence\}''\\
       & Q: \{Question\} in Bob's opinion based on the given text?\\
       \midrule 
       \multirow{8}{*}{Claude 2} & Human: \\
       & Instruction: read the given information and answer the corresponding \\
       & question. Only output the answer without other context words.\\
       & \\
       & Bob said, ``\{Evidence\}''\\
       & Q: \{Question\} in Bob's opinion based on the given text?\\
       & \\
       & Assistant:\\
       \midrule
       \multirow{7}{*}{ChatGPT \& GPT-4} & system: You are a helpful assistant.\\
       & \\
       & user: Instruction: read the given information and answer the corresponding \\
       & question. Only output the answer without other context words.\\
       & \\
       & Bob said, ``\{Evidence\}''\\
       & Q: \{Question\} in Bob's opinion based on the given text?\\
       \bottomrule
   \end{tabular}
   \caption{Opinion-based Inference Prompts for Different Models Following \cite{zhou2023context}} 
   \label{tab:bob_prompt}
\end{table*}

\begin{table*}[h!]
        \setlength{\tabcolsep}{5pt}
        \centering
        \begin{tabular}{llcccccccc}
            \toprule
            \multirow{2}{*}{Models} & \multirow{2}{*}{Method} & \multicolumn{3}{c}{Few-shot Question Only} & & \multicolumn{3}{c}{Few-shot Evidence Only} \\ 
            \cline{3-5}
            \cline{7-9}
            
             &  & EM & F1 & Entail. & & EM & F1 & Entail. \\
            \midrule
            \multirow{3}{*}{Alpaca-7B} & Closed-Book & 2.67 & 13.45 & 13.30 && 2.40 & 13.35 & 12.89\\
            & Open-Book & 23.38 & 44.94 & 60.65 && 24.56 & 46.18 & 62.87\\
            & Faithful Prompt & 30.94 & 51.88 & 63.50 && 33.06 & 54.93 & 66.21\\
            \midrule
            \multirow{3}{*}{ChatGPT} & Closed-Book & 9.81 & 25.02 & 22.03 && 9.45 & 24.78 & 21.66\\
            & Open-Book & 40.93 & 59.10 & 67.89 && 40.66 & 58.78 & 67.03\\
            & Faithful Prompt & 40.89 & 57.59 & 64.22 && 38.22 & 54.94 & 60.88\\
            \midrule
            \multirow{3}{*}{Claude 2} & Closed-Book & 6.24 & 19.49 & 22.75 && 6.11 & 19.39 & 22.70\\
            & Open-Book & 22.16 & 39.63 & 71.73 && 22.21 & 40.03 & 73.95\\
            & Faithful Prompt & 38.13 & 53.17 & 68.70 && 39.35 & 55.45 & 70.78\\
            \midrule
            \multirow{3}{*}{Palm} & Closed-Book & 11.99 & 25.23 & 21.26 && 11.99 & 25.23 & 21.26\\
            & Open-Book & 58.44 & 72.89 & 73.45 && 61.96 & 77.58 & 78.11\\
            & Faithful Prompt & 55.63 & 70.15 & 70.28 && 58.48 & 73.90 & 73.32\\
            \midrule
            \multirow{3}{*}{GPT-4} & Closed-Book & 20.76 & 38.04 & 36.14 && 20.62 & 37.98 & 35.55\\
            & Open-Book & 54.23 & 72.85 & 80.69 && 56.54 & 75.48 & 83.31\\
            & Faithful Prompt & 54.95 & 71.76 & 77.25 && 57.08 & 73.89 & 78.79\\
           
            \bottomrule
        \end{tabular}
        \caption{Few-shot result of Question-based Cat2 data and Evidence-based Cat2 data. } 
        \label{tab:3.3_cat2_single_result}
    \end{table*}

\begin{table*}[ht]
        \setlength{\tabcolsep}{5pt}
        \centering
        \begin{tabular}{lccc}
            \toprule
            Models & Start ( < 1/3 ) & Middle (1/3 - 2/3) & End (> 2/3)\\ 
            \midrule
            Alpaca-7B & 63.68 & 54.98 & 50.64 \\
            \midrule
            ChatGPT & 68.22 & 68.09	& 70.02\\
            \midrule
            Claude 2 & 71.56 & 72.04 & 70.93\\
            \midrule
            Palm & 79.24 & 80.09 & 83.54\\
            \midrule
            GPT-4 & 82.84 & 82.46 & 83.36\\
            \bottomrule
        \end{tabular}
        \caption{Few-shot entailment accuracy of Cat1 data. ``Start'', ``Middle'' and ``End'' indicates the position of the answer span in the evidence.}
        \label{tab:4.4_position_result}
    \end{table*}

\begin{table*}[h!]
        \small
        \centering
        \begin{tabular}{l|l}
            \toprule
            Cat 1 &\textbf{Question}: \\
            &  what is the baby elephants name in jungle book\\
            & \textbf{Evidence}: \\
            & Dumbo - The baby elephant who is the son of Hathi and Winifred and is a good friend of Mowgli. \\
            & He is voiced by Clint Howard in the first movie and by Jimmy Bennett in The Jungle Book 2\\
            & \textbf{Answer}: Dumbo\\
            & \textbf{GPT4 Output}: Hathi\\
            \midrule
            Cat 1 & \textbf{Question}: \\
            & who brought the idea of castles to england\\
            & \textbf{Evidence}: \\
            & Castles served a range of purposes , the most important of which were military , administrative , \\
            & and domestic . As well as defensive structures , castles were also offensive tools which could be \\
            & used as a base of operations in enemy territory . Castles were established by British rulers of \\
            & England for both defensive purposes and to pacify the country 's inhabitants . As William the \\
            & Conqueror advanced through England , he fortified key positions to secure the land he had taken . \\
            & Between 1066 and 1087 , he established 36 castles such as Warwick Castle , which he used to \\
            & guard against rebellion in the English Midlands\\
            & \textbf{Answer}: British rulers\\
            & \textbf{GPT4 Output}: William the Conqueror\\
            \midrule 
            Cat 1 & \textbf{Question}: \\
            & baga beach is in north or south goa\\
            & \textbf{Evidence}: \\
            & Baga Beach is a popular beach and tourist destination in South Goa. Baga is located at the north \\
            & end of the contiguous beach stretch that starts from Sinquerim, Candolim, leads to Calangute \\
            & and then to Baga\\
            & \textbf{Answer}: South Goa\\
            & \textbf{GPT4 Output}: North Goa\\
            \midrule
            Cat 2 Query-based & \textbf{Question}: \\
            & how long prime minister stay in office canada\\
            & \textbf{Evidence}: \\
            & The Prime Minister of Canada is appointed by the Governor General on the advice of the Prime \\
            & Minister and serves for an indefinite term, usually around 5 years. The Lieutenant Governors at \\
            & the provincial level are appointed in a similar manner and also serve for approximately 5 years. \\
            & The territories have Commissioners who are appointed by the federal cabinet and conventionally \\
            & serve for about 5 years. The Prime Minister of Canada is the head of government and chooses the \\
            & ministers that make up the Cabinet. The current Prime Minister, Justin Trudeau, remains in office \\
            & until he resigns, is dismissed, or dies.\\
            & \textbf{Answer}: until he or she resigns , is dismissed , or dies\\
            & \textbf{GPT4 Output}: Usually around 5 years\\
            \midrule
            Cat 2 Evidence-based & \textbf{Question}: \\
            & what percentage of the earth 's surface is water\\
            & \textbf{Evidence}: \\
            & Because the oceans that cover roughly 78\% of the area of the Earth reflect blue light, the Earth \\
            & appears blue from space, and is often referred to as the blue planet and the Pale Blue Dot. The \\
            & Earth's water is distributed across various sources, with oceans holding 97\% of surface water, \\
            & glaciers and polar ice caps holding 2.4\%, and other land surface water such as rivers, lakes, and \\
            & ponds holding 0.6\%. Only a small portion of water is contained in aquifers, vapor, clouds, \\
            & precipitation, biological bodies, and manufactured products. The total volume of water on Earth \\
            & is estimated to be 1.386 billion km\u00b3, with 97.5\% being saltwater and 2.5\% being freshwater. \\
            & Of the freshwater, only 0.3\% is liquid on the surface, while the rest may be present in the lower \\
            & mantle of the Earth. The United Nations Convention on the Law of the Sea defines all of the \\
            & ocean as "sea," making Earth the only known planet with liquid water on its surface. Additionally, \\
            & Earth's water distribution, including oceans, ice caps, and clouds, gives it a distinct blue \\
            & appearance when viewed from space. Approximately 97.2\% of Earth's known water is \\
            & contained within the seas, which cover more than 70\% of its surface.\\
            & \textbf{Answer}: 78\%\\
            & \textbf{GPT4 Output}: 70\%\\
            
            \bottomrule
        \end{tabular}
        \caption{Error Examples of GPT-4 under the Few-shot Setting.} 
        \label{tab:gpt4_error}
    \end{table*}

\end{document}